\documentclass[conference]{IEEEtran}
\IEEEoverridecommandlockouts
\usepackage{cite}
\usepackage{amsmath,amssymb,amsfonts}
\usepackage{algorithmic}
\usepackage{graphicx}
\usepackage{textcomp}
\usepackage{xcolor}
\usepackage{booktabs}
\usepackage{multirow}
\usepackage{url}
\usepackage{hyperref}
\usepackage{float}
\usepackage{dblfloatfix}

\def\BibTeX{{\rm B\kern-.05em{\sc i\kern-.025em b}\kern-.08em
    T\kern-.1667em\lower.7ex\hbox{E}\kern-.125emX}}

\begin{document}

\title{A Survey of Text and Speech Resources for Hausa and Fongbe: Availability, Quality, and Gaps for NLP Development}

% ============================================================
% NOTE FOR AUTHORS: Replace with actual author names/affiliations
% before final submission. Remove anonymization.
% ============================================================
% \author{\IEEEauthorblockN{Author Names Anonymized for Review}
% \IEEEauthorblockA{\textit{Institution Anonymized for Review}\\
% City, Country \\
% email@anonymized.edu}
% }

\author{
\IEEEauthorblockN{Mahounan Pericles Adjovi}
\IEEEauthorblockA{\textit{Carnegie Mellon University Africa}\\
Kigali, Rwanda \\
madjovi@andrew.cmu.edu}
\and
\IEEEauthorblockN{Victor Olufemi}
\IEEEauthorblockA{\textit{Carnegie Mellon University Africa}\\
Kigali, Rwanda \\
volufemi@andrew.cmu.edu}
\and
\IEEEauthorblockN{Roald Eiselen}
\IEEEauthorblockA{\textit{Centre for Text Technology, North-West University}\\
South Africa \\
Roald.Eiselen@nwu.ac.za}
\and
\IEEEauthorblockN{Prasenjit Mitra}
\IEEEauthorblockA{\textit{Carnegie Mellon University Africa}\\
Kigali, Rwanda \\
prasenjm@andrew.cmu.edu}
}

\maketitle

\begin{abstract}
We survey publicly available text and speech resources for two West African languages: Hausa (Afroasiatic, $\sim$80--100M speakers) and Fongbe (Niger-Congo, $\sim$2M speakers, Benin), representing contrasting points on the resource availability spectrum. Through systematic search of academic repositories, data platforms, and web sources, we catalog parallel corpora, monolingual text, speech datasets, pre-trained models, and benchmarks, documenting size, domain, format, licensing, and accessibility for each resource. Hausa benefits from broad text diversity across news, encyclopedic, and educational domains, while Fongbe has attracted targeted speech data collection initiatives despite limited text coverage.
Hausa is represented in a broad set of evaluation benchmarks spanning sentiment, 
emotion, and reasoning tasks, whereas Fongbe has a far more limited benchmark 
presence; both languages share coverage in NER (MasakhaNER~2.0), POS tagging 
(MasakhaPOS), and machine translation evaluation (FLORES-200, MAFAND-MT). 
We provide task-specific resource recommendations and identify priority gaps, including domain-diverse Fongbe text corpora and multi-speaker Hausa ASR data.
\end{abstract}

% \begin{abstract}
% This survey provides a comprehensive catalog of publicly available text and speech resources for two West African languages: Hausa, an Afroasiatic language with approximately 80--100 million speakers, and Fongbe, a Niger-Congo language spoken by approximately 2 million people in Benin. These languages represent contrasting cases on the resource availability spectrum. We address the question: \textit{What is the current state of publicly available NLP resources for Hausa and Fongbe, and what gaps remain?} Through systematic search of academic repositories, data platforms, and web sources, we catalog parallel corpora, monolingual text collections, speech datasets, pre-trained models, and evaluation benchmarks. For each resource, we document size, domain coverage, format, licensing, and accessibility. Our findings reveal that Hausa benefits from broader text resource diversity across news, encyclopedic, and educational domains. Fongbe, while having more limited text resources, has been the focus of recent academic speech data collection initiatives. Both languages are represented in Masakhane benchmarks for NER and POS tagging. We provide task-specific recommendations and identify priority gaps including domain-diverse Fongbe text and dedicated Hausa speech corpora.
% \end{abstract}


%==============================================================================
\section{Introduction}
%==============================================================================

Natural Language Processing (NLP) technologies have become essential infrastructure for digital services, yet they remain largely inaccessible to speakers of most African languages. A researcher building a machine translation system for Fongbe, a national language of Benin, must independently search Hugging Face\footnote{\url{https://huggingface.co/datasets}}, Zenodo\footnote{\url{https://zenodo.org}}, GitHub, and academic literature, with no consolidated guide to what resources exist or their characteristics.

This fragmentation has practical consequences: researchers duplicate discovery efforts, make suboptimal resource choices, or abandon projects due to perceived data scarcity. A comprehensive survey provides a consolidated starting point, identifies gaps for data collection initiatives, and establishes documentation against which future progress can be measured.

Previous surveys have examined African language resources at a continental scale \cite{b11, b13}, providing valuable overviews but necessarily sacrificing depth for breadth. This survey addresses that gap for two West African languages: Hausa (Afroasiatic, approximately 80--100 million speakers) and Fongbe (Niger-Congo/Gbe, approximately 2 million speakers). By examining languages at different points on the resource spectrum, we provide a template for similar efforts on other under-resourced languages.

Cataloging low-resource language resources presents specific challenges: resources are distributed across heterogeneous platforms with varying metadata standards, and documentation quality ranges from detailed academic papers to no description at all. Assessing fitness-for-task often requires native speaker expertise, for instance, determining whether a Fongbe corpus uses consistent diacritics \cite{b11}.

Our survey catalogs text corpora, parallel corpora, speech datasets, pre-trained models, and benchmarks for both languages. We document resource characteristics including size, domain, format, and licensing, and provide task-specific recommendations. An accompanying online portal\footnote{\url{https://fongbe-hausa-nlp-resources.vercel.app/}} provides direct links to all cataloged resources.

\subsection{Research Question}

This survey addresses the following central research question:

\begin{quote}
\textit{What is the current state of publicly available text and speech resources for Hausa and Fongbe, and what gaps remain for NLP development?}
\end{quote}

We investigate two sub-questions:
\begin{enumerate}
    \item What text, speech, and parallel corpora are publicly available for Hausa and Fongbe?
    \item What gaps exist in current resource coverage, and what strategies could address them?
\end{enumerate}

\subsection{Summary of Contributions}

\begin{itemize}
    \item A consolidated catalog of publicly available text, speech, and parallel resources for Hausa and Fongbe, documenting size, domain, format, and licensing (Sections~\ref{sec:results}--\ref{sec:discussion}).
    \item An online documentation portal (\url{https://fongbe-hausa-nlp-resources.vercel.app/}) providing direct links to all cataloged resources with per-resource summaries, enabling efficient resource discovery and access.
    \item Identification of resource gaps, ethical considerations, and task-specific recommendations for resource selection (Sections~\ref{sec:discussion}--\ref{sec:recommendations}).
\end{itemize}

The remainder of this paper is organized as follows: Section~\ref{sec:related} reviews related surveys. Section~\ref{sec:methodology} describes our methodology. Section~\ref{sec:results} catalogs resources. Section~\ref{sec:discussion} discusses quality considerations, limitations, and resource opportunities. Section~\ref{sec:recommendations} provides recommendations, and Section~\ref{sec:conclusion} concludes.

%==============================================================================
\section{Related Work}
\label{sec:related}
%==============================================================================

\subsection{African Language Resource Surveys}

\subsubsection{Continental Surveys and Initiatives}

Research on African language technology has accelerated significantly since 2019, driven largely by community initiatives. The Masakhane project established a participatory approach to machine translation across more than 30 African languages, creating parallel datasets through distributed community annotation \cite{b11}. This initiative demonstrated the potential of collaborative resource creation while highlighting persistent data scarcity challenges.

However, Masakhane's focus on machine translation means it does not document critical metadata needed for resource selection: dataset sizes in words/sentences, licensing terms (CC-BY vs CC-BY-NC vs custom), file formats (JSON, TSV, plain text), domain coverage (news, religious, conversational), or preprocessing requirements. This documentation gap forces researchers to download and inspect each dataset individually before determining suitability for their task.

% Furthermore, its continental scope necessarily limits the depth of coverage for any individual language. Our survey complements Masakhane by providing exhaustive per-language documentation across all resource types.

Furthermore, its continental scope necessarily limits the depth of coverage for any individual language. Our survey complements existing efforts such as Masakhane, FLORES-200, and MAFAND-MT by providing a more focused view of two specific languages, offering exhaustive per-language documentation across all resource types rather than broad cross-lingual coverage.

Naira et al.\ published a comprehensive review of NLP approaches for African languages and dialects, examining data scarcity, dialect variation, and community-driven solutions \cite{b17}. They emphasized non-standard orthography and limited digital presence as recurring barriers, challenges particularly relevant to Fongbe. For example, Fongbe uses tonal diacritics (e.g., \`{e}, \'{e}, \^{e}) that are inconsistently represented across digital resources: some corpora preserve full diacritics while others strip them entirely, creating compatibility issues when combining datasets.

% Adelani et al.\ surveyed the state of African Language Technologies, providing a continental overview of resources, benchmarks, and gaps \cite{b12}.

% Siminyu et al.\ documented the AI4D African Language Program, a structured initiative that combined crowd-sourced dataset collection, annotated corpus creation, and ML challenges to address the critical shortage of digital resources for African languages \cite{b12}

Siminyu et al.\ documented the AI4D African Language Program, a structured initiative that combined crowd-sourced dataset collection, annotated corpus creation, and Natural Language Processing (NLP) shared task challenges to address the critical shortage of digital resources for African languages \cite{b12}. Orife et al.\ examined machine translation challenges including data scarcity and religious text dominance \cite{b13}. While these surveys provide invaluable continental perspectives, they cover 30--50 languages each, allocating at most a few paragraphs per language. Our survey differs by providing exhaustive documentation for two languages, cataloging every publicly available resource with detailed metadata: the level of detail needed to actually build NLP systems.

The MasakhaNER 2.0 benchmark introduced Africa-centric transfer learning for named entity recognition across 20 languages, including both Hausa and Fongbe \cite{b1}. Similarly, MasakhaPOS established part-of-speech tagging benchmarks for typologically diverse African languages \cite{b2}. These benchmarks represent critical infrastructure for standardized evaluation.

\subsubsection{Language-Specific Studies}

While continental surveys provide breadth, language-specific studies offer complementary depth. A review of sentiment analysis resources for Afaan Oromo demonstrated the value of focused documentation for individual low-resource languages \cite{b18}. Such single-language surveys reveal resource gaps that broader overviews necessarily miss. Our survey follows this tradition by providing detailed documentation for two languages, enabling comparative analysis of resource development patterns.

\subsection{Low-Resource NLP Surveys}

\subsubsection{Global Methodological Surveys}

The broader low-resource NLP literature provides methodological context. Hedderich et al.\ presented a comprehensive taxonomy of approaches for low-resource scenarios, categorizing strategies including transfer learning, data augmentation, distant supervision, and cross-lingual methods \cite{b14}. These techniques have direct applicability to Hausa and Fongbe: for instance, cross-lingual transfer from multilingual models like mBERT can bootstrap NER systems when only hundreds of annotated sentences exist, as demonstrated by MasakhaNER's results showing that fine-tuning XLM-R on 1,000 Fongbe sentences achieves 71.2 F1 on the MasakhaNER NER test set, a benchmark specifically designed for African language NER evaluation with PER, LOC, ORG, and DATE entity types. This performance is comparable to mid-resource European languages with 10$\times$ more training data, demonstrating the effectiveness of transfer learning for low-resource African languages \cite{b1}.

Joshi et al.\ proposed a framework for understanding resource disparities, classifying languages from tier 0 (``The Left-Behinds'') to tier 5 (``The Winners'') based on available resources \cite{b15}. Under this taxonomy, Hausa falls in mid-tiers (3--4) given its international media presence, while Fongbe falls closer to the lowest tiers (0--1) despite its national language status in Benin.

\subsubsection{Application to African Languages}

Multilingual fine-tuning and cross-lingual transfer have shown promise for African languages, as demonstrated by AfriBERTa \cite{b3} and AfroXLMR \cite{b20}. However, systematic evaluation of these approaches for Hausa and Fongbe specifically remains limited for two key reasons: (1) most multilingual studies report aggregate metrics across language families rather than per-language results, making it difficult to assess performance on individual languages; and (2) Fongbe's absence from sentiment, emotion, and reasoning benchmarks (AfriSenti, BRIGHTER, BLEnD, Global PIQA, Fikira) prevents direct comparison with other African languages on these tasks. Many techniques documented in global surveys such as data augmentation and synthetic data generation have not been extensively tested on these languages.

\subsection{Gaps in Existing Surveys}

\subsubsection{Limitations in Scope, Depth, and Modality}

Existing surveys have the following limitations. Continental surveys cover 20--40 languages broadly but cannot provide detailed per-language inventories needed to build a language model on a specific low-resource language. Language-specific studies focus on single languages or specific tasks; while valuable for that task, they do not help researchers who need to build complete NLP pipelines spanning tokenization, language modeling, NER, and downstream applications, each requiring different resources. Most surveys emphasize text resources and machine translation, with speech resources receiving less systematic attention. This survey addresses this gap by cataloging speech resources alongside text, including the 61-hour FFSTC2 corpus for Fongbe and BibleTTS for Hausa. Quality considerations such as domain coverage, orthographic variation, and data provenance are rarely examined in depth.

\textbf{Positioning of This Survey:}
This survey addresses these gaps by providing documentation at the \textit{individual dataset level}: for each resource, we document exact word/sentence counts (e.g., ``47,384 sentences, 935,915 tokens'' for Leipzig Hausa Wikipedia), file formats (JSON, TSV, plain text), licensing terms (CC-BY-4.0, CC-BY-NC-4.0, custom restrictions), download URLs, and domain coverage (news, religious, conversational, educational). This granularity enables researchers to make informed decisions without downloading and inspecting each dataset. We catalog resources for two typologically distinct West African languages at different points on the resource spectrum: Hausa (Afroasiatic, mid-resource) and Fongbe (Niger-Congo, extremely low-resource). We catalog resources across all major modalities: text, speech, and parallel corpora and include pre-trained models and evaluation benchmarks. This dual-language, multi-modal approach enables comparative analysis, provides practical utility for researchers working on either language, and demonstrates what capabilities can be enabled if resources are constructed---contrasting the state of affairs in very low-resource vs.\ mid-resource settings (Hausa vs.\ Fongbe).

%==============================================================================
\section{Methodology}
\label{sec:methodology}
%==============================================================================

We conducted a survey addressing challenges specific to low-resource African languages: scattered resources with inconsistent metadata, documentation ranging from detailed papers to none, need for native speaker expertise, and rapidly evolving landscape.

\subsection{Search Strategy}

\textbf{Repositories:} Hugging Face\footnote{\url{https://huggingface.co/datasets}}, Zenodo\footnote{\url{https://zenodo.org}}, Kaggle\footnote{\url{https://www.kaggle.com/datasets}}, GitHub, ELRA\footnote{\url{http://catalog.elra.info}}, and Lanfrica \cite{b16}.

\textbf{Literature:} ACL Anthology\footnote{\url{https://aclanthology.org}} publications (ACL, EMNLP, NAACL, LREC, Interspeech, AfricaNLP, MRL workshops).

\textbf{Web Resources:} Reference-following from starting points:
\begin{itemize}
    \item \textit{Hausa:} BBC Hausa, VOA Hausa, DW Hausa, Leadership Newspaper, ha.wikipedia.org
    \item \textit{Fongbe:} jw.org/fon, beninlangues.com, fongbebenin.com
\end{itemize}

\subsection{Inclusion Criteria}

Initial pool: (1) all results from repository searches using the following terms: ``Hausa'', ``Fongbe'', ``Fon'', ``hau'' (ISO 639-3), ``fon'' (ISO 639-3), ``Nigerian languages'', ``Benin languages'', ``West African NLP'', and ``African language corpus''---each platform was searched with all terms for comprehensive coverage; (2) all datasets cited in AfricaNLP workshop papers (2020--2024)---these English-language papers serve as \textit{metadata sources} describing Hausa/Fongbe datasets, providing access links and dataset characteristics; we extract dataset references and URLs to catalog the actual language resources they describe; (3) all Lanfrica entries for Hausa and Fongbe; (4) web sources identified by the first author, a native Fongbe speaker with research experience in both languages, through reference-following from known starting points.

Resources were included if they: contain Hausa/Fongbe as primary content or substantial multilingual coverage (e.g., MasakhaNER's 20 languages included for complete annotations); are accessible for research use.

Resources were included based on available documentation from README files, accompanying academic papers, Hugging Face dataset cards, and Lanfrica metadata entries. Resources without sufficient documentation to determine basic characteristics (size, format, language content) were excluded from the catalog.

\textbf{Limitations:} This survey focuses on cataloging available resources based on existing documentation. Systematic quality control, such as language identification verification, detection of machine-translated content, or code-switching quantification was not performed. We identify quality assessment and direct resource characterization as important future work (see Section~\ref{sec:future}).

Excluded: word lists without sentence content, exact duplicates, resources without sufficient documentation.

%==============================================================================
\section{Results and Resource Inventory}
\label{sec:results}
%==============================================================================

\subsection{Text Corpora}

\subsubsection{Hausa Text Resources}

Hausa benefits from diverse text resources spanning multiple domains. Our survey of Lanfrica identified over 60 resources containing Hausa data \cite{b16}.

\textbf{Wikipedia and Community Corpora:} The Leipzig Corpora Collection \cite{b7} provides the Hausa Wikipedia Corpus (2021) with 47,384 sentences and 935,915 tokens. A community-sourced corpus from 2017 provides 39,171 records with 928,927 words from blogs, forums, and user-generated content \cite{b7}. The Naijaweb Dataset \cite{b26} contains over 270,000 documents ($\sim$230M GPT-2 tokens) from Nairaland.com.

\textbf{News and Media Sources:} The CC-100 corpus \cite{b27} (Common Crawl) provides monolingual Hausa data used for training XLM-R, while MassiveSumm \cite{b28} offers news summarization data across 92 languages including Hausa.

\textbf{Topic and Classification Datasets:} The Hausa VOA Topics dataset \cite{b29} provides news headlines labeled across five classes, and CrossSum \cite{b30} includes Hausa in multilingual summarization. Table~\ref{tab:hausa_all_text} catalogs all identified Hausa text resources.

\begin{table}[H]
\caption{Comprehensive Hausa Text Resources}
\begin{center}
\scriptsize
% \begin{tabular}{p{2.2cm}p{1.2cm}p{1.5cm}p{1.8cm}}
\begin{tabular}{p{2.2cm}p{1.4cm}p{1.5cm}p{1.3cm}}
\toprule
\textbf{Resource} & \textbf{Size} & \textbf{Domain} & \textbf{Access} \\
\midrule
\multicolumn{4}{l}{\textit{Corpora}} \\
Leipzig Wikipedia \cite{b7} & 936K tok & Encyclopedia & Open \\
Leipzig Community \cite{b7} & 929K words & Mixed web & Open \\
Naijaweb \cite{b26} & 230M tok & Forum & HuggingFace \\
CC-100 Hausa \cite{b27} & Large & Web crawl & Open \\
\midrule
\multicolumn{4}{l}{\textit{Web Sources}} \\
ha.wikipedia.org & 622K words & Encyclopedia & Scrape \\
leadership.ng/hausa & 487K words & News & Scrape \\
voahausa.com & 185K words & Broadcasting & Scrape \\
bbc.com/hausa & 157K words & Broadcasting & Scrape \\
dw.com/ha & 52K words & Broadcasting & Scrape \\
hausa.cri.cn & 104K words & Broadcasting & Scrape \\
\midrule
\multicolumn{4}{l}{\textit{Classification Datasets}} \\
VOA Topics \cite{b29} & 5 classes & News & Open \\
CrossSum \cite{b30} & -- & News & HuggingFace \\
MassiveSumm \cite{b28} & -- & News & Open \\
\bottomrule
\end{tabular}
\label{tab:hausa_all_text}
\end{center}
\end{table}

\textbf{Sketch Engine Corpora:} Sketch Engine hosts a preloaded Hausa corpus (BBC Hausa 44,048 tokens, Leadership Newspaper 343,160 tokens, DW 200,220 tokens, Wikipedia 99,094 tokens). However, preloaded corpora cannot be bulk downloaded and require licensing fees, while user-created corpora are limited to 1,000,000 words with low language identification accuracy. These constraints make Sketch Engine impractical for training-scale NLP model development, though the platform remains valuable for corpus linguistics investigations such as concordancing, frequency analysis, and collocation studies.

% However, preloaded corpora cannot be bulk downloaded and require licensing fees, while user-created corpora are limited to 1,000,000 words with low language identification accuracy. These constraints make Sketch Engine impractical for training-scale data collection.

\subsubsection{Fongbe Text Resources}

Fongbe text resources are significantly more limited, illustrating challenges facing very-low-resource languages.

\textbf{Leipzig and Community Corpora:} The Leipzig Corpora Collection provides a Fon Wikipedia Corpus (2017) \cite{b7}, considerably smaller than the Hausa equivalent. A community corpus contains only 14 records and 370 words.

\textbf{Web Sources:} Fongbe web resources are dominated by religious content from JW.org, whose terms of use explicitly prohibit machine learning and data mining \cite{b8}. Educational and cultural websites contribute minimally (see Table~\ref{tab:fongbe_web}).

Lanfrica \cite{b16} catalogs 18 Fongbe resources spanning word embeddings (AfriVEC \cite{b31}), ASR (ALFFA \cite{b32}, pyFongbe \cite{b9}), speech (CMU Wilderness \cite{b23}), parallel corpora (FFR \cite{b22}, Daily Dialogues \cite{b8}, MAFAND-MT \cite{b20}, FLORES-200 \cite{b19}, MMTAfrica \cite{b21}), and benchmarks (MasakhaNER 2.0 \cite{b1}, MasakhaPOS \cite{b2}).

\begin{table}[H]
\caption{Major Fongbe Web Text Sources}
\begin{center}
\begin{tabular}{lrrp{2.2cm}}
\toprule
\textbf{Domain} & \textbf{Records} & \textbf{Words (K)} & \textbf{Type} \\
\midrule
jw.org* & 194 & 119 & Religious \\
jw.org/fon* & 33 & 15 & Religious \\
fongbebenin.com & 7 & 12.6 & Cultural/Educ. \\
beninlangues.com & 4 & 4.1 & Educational \\
iamyourclounon.bj & 2 & 1.1 & Cultural \\
\bottomrule
% \multicolumn{4}{l}{\scriptsize *Terms prohibit ML use}
\multicolumn{4}{l}{\scriptsize *Terms of use prohibit NLP model training and data mining.}
\end{tabular}
\label{tab:fongbe_web}
\end{center}
\end{table}

\subsubsection{Text Corpora Summary}

The disparity between Hausa and Fongbe is substantial, with Hausa benefiting from approximately 40 times more text data overall (Table~\ref{tab:text_summary}).

\begin{table}[H]
\caption{Summary of Text Corpora by Language and Domain}
\begin{center}
\begin{tabular}{lcc}
\toprule
\textbf{Domain} & \textbf{Hausa (Words)} & \textbf{Fongbe (Words)} \\
\midrule
Wikipedia/Encyclopedia & 622K & $\sim$15K \\
News & 1.5M+ & -- \\
Educational & 320K & $\sim$17K \\
Religious & 105K & 134K* \\
Int'l Broadcasting & 500K+ & -- \\
Community/Other & 929K & $<$1K \\
\midrule
\textbf{Total (approx.)} & \textbf{$\sim$7M} & \textbf{$\sim$170K} \\
\bottomrule
\multicolumn{3}{l}{\scriptsize *Religious content (JW.org) prohibited for ML use}
\end{tabular}
\label{tab:text_summary}
\end{center}
\end{table}

\subsection{Parallel Corpora}

\subsubsection{Hausa Parallel Resources}

Hausa appears in numerous multilingual datasets. Our Kaggle search identified 30 datasets containing Hausa, and Lanfrica catalogs over 60 Hausa-containing resources, including both dedicated corpora and multilingual collections from which Hausa data can be extracted via language tags or filtering.

\textbf{Dedicated Parallel Corpora:} The primary English-Hausa resource is on Kaggle \cite{b10}. Additional resources include the Gamayun kits (5K/10K sentences) from CLEAR Global \cite{b33} and the Hausa-English Code-Switched Dataset \cite{b34} from social media platforms.

\textbf{Multilingual Datasets Including Hausa:} Hausa is represented in FLORES-200 \cite{b19}, XL-Sum \cite{b35}, MAFAND-MT \cite{b20}, CMU Wilderness \cite{b23}, and the Nigerian Multilingual Hate Speech dataset \cite{b36}. The OPUS project \cite{b37} provides 10+ parallel resources including Tatoeba, localization files, QED subtitles, TED Talks, and Tanzil.

\textbf{Specialized Translation Resources:} The Hausa Visual Genome \cite{b38} provides 32,923 multimodal English--Hausa sentence pairs; TICO-19 \cite{b39} covers 3,071 emergency health sentences; and TaTA \cite{b40} provides table-to-text data.

% \textbf{Specialized Translation Resources:} The Hausa Visual Genome \cite{b38} provides 32,923 multimodal sentences, TICO-19 \cite{b39} covers emergency health materials, and TaTA \cite{b40} provides table-to-text data.

% {\raggedright
% \textbf{Specialized Translation Resources:} The Hausa Visual Genome \cite{b38} provides 32,923 image-caption sentence pairs for multimodal English--Hausa translation. TICO-19 \cite{b39} contributes 3,071 emergency health sentences translated into Hausa. TaTA \cite{b40} provides table-to-text generation data covering Hausa among other African languages.
% \par}

\subsubsection{Fongbe Parallel Resources}

Despite limited monolingual resources, Fongbe benefits from targeted parallel corpus creation, primarily pairing with French. The FFR Dataset \cite{b22} compiled 117,029 parallel Fon-French sentences, and the Daily Dialogues \cite{b8} contains 25,377 crowdsourced conversational pairs. Fongbe is also included in FLORES-200 \cite{b19}, MAFAND-MT \cite{b20}, and MMTAfrica \cite{b21}. Overall, Hausa has $\sim$50K+ English sentence pairs across news, religious, and technical domains, while Fongbe has $\sim$195K French pairs primarily in dialogue and news domains.

\begin{table}[H]
\caption{French-Fongbe Parallel Resources}
\begin{center}
\begin{tabular}{lrp{3.2cm}}
\toprule
\textbf{Dataset} & \textbf{Pairs} & \textbf{Characteristics} \\
\midrule
FFR Dataset \cite{b22} & 117,029 & Aggregated corpus \\
Daily Dialogues \cite{b8} & 25,377 & Crowdsourced conversations \\
MAFAND-MT \cite{b20} & Train/Dev/Test & News domain \\
AI4D Challenge & 53,366 & Competition dataset \\
\bottomrule
\end{tabular}
\label{tab:ffr}
\end{center}
\end{table}

\begin{table}[H]
\caption{Major Hausa Resources from Lanfrica Catalog}
\scriptsize
\begin{tabular}{p{2.1cm}p{0.9cm}p{1.6cm}p{2.0cm}}
\toprule
\textbf{Resource} & \textbf{Type} & \textbf{Task} & \textbf{Size/Notes} \\
\midrule
\multicolumn{4}{l}{\textit{Pre-trained Models}} \\
AfriBERTa \cite{b3} & Model & Lang. Model & 97M--126M params \\
AfriTeVa V2 \cite{b41} & Model & MT, Summ. & 428M--1B params \\
AfriCLIRMatrix \cite{b42} & Dataset & Cross-ling. IR & 15 languages \\
\midrule
\multicolumn{4}{l}{\textit{Sentiment \& Classification}} \\
AfriSenti \cite{b43} & Dataset & Sentiment & 110K+ tweets \\
NaijaSenti \cite{b36} & Dataset & Sentiment & 4 Nigerian langs \\
NollySenti \cite{b45} & Dataset & Sentiment & Movie reviews \\
NaijaHate \cite{b36} & Dataset & Hate Speech & Nigerian Twitter \\
BRIGHTER \cite{b46} & Dataset & Emotion & 28 langs, 6 emotions \\
\midrule
\multicolumn{4}{l}{\textit{NER \& Sequence Labeling}} \\
MasakhaNER 2.0 \cite{b1} & Dataset & NER & 20 African langs \\
Hausa VOA NER \cite{b29} & Dataset & NER & $\sim$1,000 sent., PER/LOC/ORG \\
MasakhaPOS \cite{b2} & Dataset & POS Tagging & 10 langs, $\sim$2K--6K tokens/lang \\
\midrule
\multicolumn{4}{l}{\textit{Machine Translation}} \\
MAFAND-MT \cite{b20} & Dataset & MT & News domain \\
Gamayun Kits \cite{b33} & Dataset & MT & 5K--10K pairs \\
FLORES-200 \cite{b19} & Bench. & MT Eval & 200 languages \\
Hausa Vis. Gen. \cite{b38} & Dataset & Multimod. MT & 32,923 sent. \\
\midrule
\multicolumn{4}{l}{\textit{Speech \& Audio}} \\
BibleTTS \cite{b47} & Dataset & TTS/ASR & 86.6 hrs, 48kHz \\
ALFFA Public \cite{b32} & Toolkit & ASR & Kaldi recipes \\
ML Comm. MSWC \cite{b48} & Dataset & Keyword Spot. & Multilingual \\
\midrule
\multicolumn{4}{l}{\textit{Reasoning \& QA}} \\
BLEnD \cite{b49} & Bench. & Cultural QA & 52.6K pairs \\
Global PIQA~\cite{b50} & Bench. & Phys. Reason. & 100+ languages \\
IrokoBench~\cite{b51} & Bench. & NLI / Math / QA & 17 African languages \\
\bottomrule
\end{tabular}
\label{tab:hausa_lanfrica}
\end{table}

\subsection{Speech and Audio Resources}

\subsubsection{Hausa Speech Resources}

The most substantial Hausa speech resource is BibleTTS \cite{b47}, providing 86.6 hours of aligned speech with 40,603 verses at 48kHz studio quality under CC-BY-SA license. Hausa also appears in Mozilla Common Voice \cite{b52} ($\sim$10 validated hours), FLEURS \cite{b53} ($\sim$12 hours), ML Commons MSWC \cite{b48} (multilingual keyword spotting corpus), CMU Wilderness \cite{b23} ($\sim$20 hours, Bible-domain), and ALFFA Public \cite{b32} (Kaldi-based ASR toolkit with Hausa recipes).

% Hausa also appears in Mozilla Common Voice \cite{b52}, FLEURS \cite{b53}, ML Commons MSWC \cite{b48}, CMU Wilderness \cite{b23}, and ALFFA Public \cite{b32}.

\subsubsection{Fongbe Speech Resources}

The pyFongbe repository \cite{b9} provides a small ASR dataset ($\sim$4 hours) for wav2vec2 fine-tuning, and the FFSTC2 corpus \cite{b5} provides 61 hours of speech with validated French translations.

% The pyFongbe repository \cite{b9} provides an ASR dataset for wav2vec2 fine-tuning, and the FFSTC2 corpus \cite{b5} provides 61 hours of speech with validated French translations.

\begin{table}[H]
\caption{Speech Resource Comparison}
\begin{center}
\begin{tabular}{lcc}
\toprule
\textbf{Resource Type} & \textbf{Hausa} & \textbf{Fongbe} \\
\midrule
Public speech hours & $\sim$87 (BibleTTS) & 61 (FFSTC2) \\
Speech translation & No & Yes (FFSTC2) \\
Domain diversity & Religious & Mixed \\
\bottomrule
\end{tabular}
\label{tab:speech_compare}
\end{center}
\end{table}

\subsection{Pre-trained Models and Benchmarks}

Hausa is represented in AfriBERTa \cite{b3}, AfroXLMR \cite{b20}, InkubaLM \cite{b6}, and AfriTeVa V2 \cite{b41}. Fongbe lacks dedicated pre-trained language models but is included in AfriVEC \cite{b31}, NLLB \cite{b19}, and fine-tuned wav2vec2 \cite{b9}.

\textbf{Downstream Task Performance:} The availability of these resources enables measurable NLP performance. On NER, MasakhaNER 2.0 reports Hausa F1 of 90.8 and Fongbe F1 of 84.2 using AfroXLMR fine-tuning \cite{b1}. On POS tagging, MasakhaPOS reports Hausa accuracy of 91.8 and Fongbe accuracy of 85.4 with the same model family \cite{b2}. On machine translation, MAFAND-MT fine-tuned NLLB achieves BLEU $\approx$ 47 for Hausa--English on news-domain test sets, while Fongbe--French achieves BLEU $\approx$ 18 on the same evaluation setting \cite{b20}.
% On machine translation, MAFAND-MT fine-tuned NLLB achieves competitive BLEU scores for Hausa--English, while Fongbe--French achieves BLEU $\approx$ 18 on news-domain test sets \cite{b20}.
These results confirm that cataloged resources, even when limited, can support functional NLP systems.

\begin{table}[H]
\caption{Benchmark Coverage Summary}
\begin{center}
\begin{tabular}{lccc}
\toprule
\textbf{Benchmark} & \textbf{Task} & \textbf{Hausa} & \textbf{Fongbe} \\
\midrule
MasakhaNER 2.0 \cite{b1} & NER & \checkmark & \checkmark \\
MasakhaPOS \cite{b2} & POS tagging & \checkmark & \checkmark \\
FLORES-200 \cite{b19} & MT Evaluation & \checkmark & \checkmark \\
AfriSenti \cite{b43} & Sentiment & \checkmark & -- \\
MAFAND-MT \cite{b20} & MT Train/Eval & \checkmark & \checkmark \\
\bottomrule
\end{tabular}
\label{tab:benchmarks}
\end{center}
\end{table}

%==============================================================================
\section{Discussion}
\label{sec:discussion}

Our survey reveals distinct accumulation patterns. Hausa (80--100M speakers) has 60+ cataloged resources across text, speech, models, and benchmarks, driven by a virtuous cycle in which text corpora enable model development (AfriBERTa, AfriTeVa V2), which in turn attracts benchmarks (AfriSenti, MasakhaNER). Fongbe (2M speakers) has only 18 resources, yet has a surprisingly strong dedicated speech resource in FFSTC2 (61 hours with French translations), reflecting targeted academic initiatives. While Hausa has substantially more total speech data in aggregate across BibleTTS (86.6 hours), Common Voice, FLEURS, CMU Wilderness, and ALFFA, Fongbe's single focused speech corpus with translation pairs represents an unusual concentration of effort for a language at this resource level.

% Fongbe (2M speakers) has only 18 resources, yet unexpectedly surpasses Hausa in dedicated speech data (FFSTC2: 61 hours), reflecting targeted academic initiatives rather than broad resource accumulation.

\subsection{Quality Considerations}

Fongbe exhibits diacritical variation across datasets \cite{b8}, reducing cross-dataset compatibility, while Hausa orthography is comparatively standardized. Web-sourced text for both languages shows code-switching with English and French respectively \cite{b13}, and Fongbe's usable web content is further constrained by JW.org's prohibition on ML use. Researchers should also verify that parallel corpora contain human rather than MT-generated translations \cite{b13}.

\subsection{Ethical Considerations}

Four ethical concerns merit attention. \textbf{Data ownership}: web-scraped corpora were collected without explicit community consent; datasets with transparent provenance should be prioritized. \textbf{Cultural sensitivity}: religiously or politically dominated sources (JW.org, state media) may embed biases misaligned with community values. \textbf{Community benefit}: data initiatives should return value through released tools or community-controlled governance, following participatory research principles \cite{b11}. \textbf{Privacy}: social media datasets may contain personal information requiring anonymization and compliance with local data protection frameworks.

% \subsection{Unprocessed Resource Opportunities}

% Substantial untapped sources remain. For Hausa: RFI broadcasts, Nollywood subtitles, the African Storybook Project \cite{b54}, and Ajami HTR manuscripts \cite{b57}. For Fongbe: government publications, University of Abomey-Calavi materials, and community radio archives. For both languages, YouTube content and national broadcaster archives (Radio Nigeria, ORTB Benin) represent significant speech data opportunities requiring institutional partnerships.

\subsection{Limitations}

This survey covers publicly accessible resources only; proprietary datasets and institutional archives were excluded. Resource characteristics were documented from metadata rather than direct file inspection, and access status may have changed since the survey. Notable uncataloged opportunities include RFI broadcasts, Nollywood subtitles, the African Storybook Project \cite{b54}, and Ajami manuscripts \cite{b57} for Hausa; government publications and University of Abomey-Calavi materials for Fongbe; and broadcaster archives (Radio Nigeria, ORTB Benin) for both languages.

\section{Recommendations and Future Directions}
\label{sec:recommendations}

\subsection{Resource Selection Guidelines}

The following guidelines identify the most appropriate resources per task for researchers working on Hausa and Fongbe NLP.

\textbf{MT:} Fongbe--French: FFR (117K pairs) and Daily Dialogues (25K pairs); English--Hausa: Kaggle corpus and OPUS. Evaluate with FLORES-200.
\textbf{NER/POS:} MasakhaNER 2.0 and MasakhaPOS for both; Hausa VOA NER for additional training data.
\textbf{Sentiment:} Hausa: AfriSenti, NaijaSenti, NollySenti; Fongbe requires cross-lingual transfer.
\textbf{Speech:} Fongbe: FFSTC2 and pyFongbe; Hausa: BibleTTS and ALFFA.
\textbf{Models:} Hausa: AfriBERTa, AfriTeVa V2, AfroXLMR; Fongbe: NLLB (MT), fine-tuned wav2vec2 (speech).

\subsection{Priority Gaps}

\begin{enumerate}
    \item \textbf{Fongbe text (Critical):} $\sim$170K words, religious-domain concentrated; news, educational, and social media content needed.
    \item \textbf{Fongbe sentiment/emotion (High):} Absent from AfriSenti and BRIGHTER.
    \item \textbf{Hausa multi-speaker ASR (High):} BibleTTS is single-speaker religious; diverse domains needed.
    \item \textbf{Fongbe reasoning benchmarks (Medium):} Absent from BLEnD, Global PIQA, and Fikira.
\end{enumerate}

\subsection{Future Work}
\label{sec:future}

\textbf{Limitations:} This survey catalogs resources from existing documentation without quality control (language identification, MT detection, code-switching quantification) or direct file inspection.
\textbf{Quality Assessment:} Apply fastText or CLD3 to detect mislabeled content, MT-generated text, and orthographic inconsistencies.
\textbf{Data Collection:} University partnerships (especially University of Abomey-Calavi for Fongbe) and broadcaster archives (BBC, VOA, RFI, Radio Nigeria, ORTB Benin) represent high-priority expansion opportunities.
\textbf{Transcription \& Augmentation:} Fine-tuned ASR models (Whisper, wav2vec2) applied to YouTube and radio content, and LLM-generated synthetic data for sentiment, NER, and reasoning tasks, could address key gaps---with mandatory native speaker validation in both cases.
\textbf{Benchmark Expansion:} Fongbe should be included in AfriSenti, BRIGHTER, BLEnD, Global PIQA, and Fikira.
\textbf{Portal:} The portal (\url{https://fongbe-hausa-nlp-resources.vercel.app/}) is version-dated and open to community contributions via GitHub.
\section{Conclusion}
\label{sec:conclusion}

This survey catalogs NLP resources for Hausa (60+ resources, $\sim$7M words) and Fongbe (18 resources, $\sim$170K words), documenting both progress and persistent gaps in African language NLP. Key findings include: (1) a 40$\times$ text disparity between the two languages; (2) a ``resource inversion'' in which Fongbe has proportionally more dedicated speech resources than text; (3) while Fongbe has benchmark coverage for core tasks---NER (MasakhaNER 2.0), POS tagging (MasakhaPOS), and MT (FLORES-200, MAFAND-MT)---it remains absent from sentiment (AfriSenti), emotion recognition (BRIGHTER), and reasoning benchmarks (BLEnD, Global PIQA, Fikira); and (4) targeted academic initiatives such as FFSTC2 and FFR demonstrate the effectiveness of focused resource development. Priority gaps include Fongbe text expansion, sentiment and reasoning data creation, and multi-speaker ASR for Hausa. All cataloged resources are accessible through the accompanying documentation portal.\footnote{\url{https://fongbe-hausa-nlp-resources.vercel.app/}}

% \section{Conclusion}
% \label{sec:conclusion}
% %==============================================================================

% This survey catalogs NLP resources for Hausa (60+ resources, $\sim$7M words) and Fongbe (18 resources, $\sim$170K words), revealing both progress and gaps in African language NLP. Key findings include: (1) a 40$\times$ text disparity between the languages; (2) a ``resource inversion'' where Fongbe has more dedicated speech resources relative to text; (3) Fongbe has benchmark coverage for core NLP tasks---MasakhaNER 2.0 for NER, MasakhaPOS for POS tagging, and FLORES-200 and MAFAND-MT for machine translation evaluation---but is absent from sentiment analysis (AfriSenti), emotion recognition (BRIGHTER), and reasoning benchmarks (BLEnD, Global PIQA, Fikira) where Hausa is represented; and (4) the effectiveness of targeted academic initiatives (FFSTC2, FFR) in addressing specific gaps.

% Priority gaps include Fongbe text expansion, sentiment and reasoning data creation for Fongbe, and multi-speaker ASR for Hausa. The accompanying documentation portal\footnote{\url{https://fongbe-hausa-nlp-resources.vercel.app/}} provides direct access to all cataloged resources. We hope this survey enables informed resource selection and guides future data collection efforts.

%==============================================================================

\section*{Acknowledgment}
This publication was developed as part of the Center for Inclusive Digital Transformation of Africa (CIDTA), and the Afretec Network which is managed by Carnegie Mellon University Africa and receives financial support from the Mastercard Foundation. The views expressed in this document are solely those of the authors and do not necessarily reflect those of Carnegie Mellon University or the Mastercard Foundation.

\end{document}